\documentclass[runningheads]{llncs}
\usepackage{graphicx}
\usepackage{graphbox}
\usepackage{amsmath,amssymb} 
\usepackage{color}
\usepackage{algorithmic}
\usepackage{algorithm}
\usepackage{commath}
\usepackage{tabularx}
\usepackage{array,mwe}
\usepackage{mathtools}
\begin{document}
\pagestyle{headings}
\mainmatter
\def\ACCV18SubNumber{16421312321}  

\titlerunning{Editable Generative Adversarial Network}
\authorrunning{Kyungjune Baek, Duhyeon Bang and Hyunjung Shim}

\title{Editable Generative Adversarial Networks: Generating and Editing Faces Simultaneously}  
\author{Kyungjune Baek, Duhyeon Bang and Hyunjung Shim}

\institute{School of Integrated Technology, Yonsei University, South Korea}

\maketitle

\begin{abstract}
We propose a novel framework for simultaneously generating and manipulating the face images with desired attributes. While the state-of-the-art attribute editing technique has achieved the impressive performance for creating realistic attribute effects, they only address the image editing problem, using the input image as the condition of model. Recently, several studies attempt to tackle both novel face generation and attribute editing problem using a single solution. However, their image quality is still unsatisfactory. Our goal is to develop a single unified model that can simultaneously create and edit high quality face images with desired attributes. A key idea of our work is that we decompose the image into the latent and attribute vector in low dimensional representation, and then utilize the GAN framework for mapping the low dimensional representation to the image. In this way, we can address both the generation and editing problem by learning the generator. 
For qualitative and quantitative evaluations, the proposed algorithm outperforms recent algorithms addressing the same problem. Also, we show that our model can achieve the competitive performance with the state-of-the-art attribute editing technique in terms of attribute editing quality. 
\end{abstract}

\section{Introduction}
Facial attribute manipulation has been an important problem in computer vision and graphics field for past several decades. Traditional techniques \cite{blanz1999morphable,blanz2003face,paysan20093d} develop geometric and radiometric facial models, and reparameterize their model parameters for representing various facial attributes. (e.g., expression, race, gender, hair and eye colors, etc.) Specifically, they derive the compact representation of attribute parameters by a linear combination of facial parameters. Such a linear model allows real time face rendering and manipulation. However, due to the simplicity of model, it is limited to represent the diversity of attributes from individuals.

Recently, there are interesting approaches to directly generating faces of realistic attributes without constructing facial models. The most representative models include \cite{he2017arbitrary,choi2017stargan,sun2018mask}, which utilizes non-linear deep generative adversarial networks (GAN) for face generation. Owing to the impressive visual quality, GANs receive increasing attentions from various researchers and practitioners. GAN aims to reproduce complex data distribution based on adversarial learning between two networks, a generator and a discriminator. The original GAN and its variants are fully unsupervised in that they do not require labeled data to learn the data generation process. As a drawback, they are incapable of controlling various attributes of data during generation. To address this issue, the conditional information such as image labels \cite{mirza2014conditional,odena2016conditional} is utilized for developing GANs to control attributes during image synthesis in a supervised fashion. Likewise,   He et al., Choi et al. and Sun et al. \cite{he2017arbitrary,choi2017stargan,sun2018mask} achieve the facial attribute editing by adopting the conditional information. 

For the successful facial attribute editing, existing approaches pursue two objectives; the facial identity after manipulation should be retained, and the effect of desired attribute should be clearly observable in the resultant face. In order to preserve the facial identity, the encoder-decoder architecture is commonly adopted, allowing to reconstruct the input face. For controlling the effect of attributes, attribute classifiers are popularly employed, where they provide the condition of generation process. As a result, existing techniques maintain the quality of input faces while generating realistic attributes. 

Although the encoder-decoder architecture is an effective tool for the identity preservation, this always requires the input face for attribute editing, thus incapable of generating new faces. VAE/GAN \cite{larsen2015autoencoding} addresses this problem by combining the VAE with GAN architecture, and achieves both the data generation and reconstruction/editing simultaneously. Unfortunately, their visual quality and reconstruction accuracy are much worse than other facial editing approaches because the variational inference degrades the quality of reconstruction/editing as well as that of generation \cite{goodfellow2016nips}. Later, to reconstruct the image for editing,
Perarnau et al. \cite{perarnau2016invertible} proposes a new framework that combines the conditional GAN and two encoders without variational inference. The first encoder maps the image generated by the generator into the latent vector. The second encoder is used to map the image to the condition vector. To train the IcGAN, the conditional GAN is first trained and then two encoders are later updated by fixing the conditional GAN for stabilizing the training process.

Our goal is similar to VAE/GAN and IcGAN in that we aim to modify facial attributes of input faces as well as generate new faces with desirable attributes. 
We focus on improving the quality of generation and reconstruction/editing compared to VAE/GAN and IcGAN. Moreover, we formulate an end-to-end network model as opposed to IcGAN. To this end, our model is developed based on the standard GAN architecture, where the generator learns an unidirectional mapping from the input prior distribution $P_z$ to the data distribution $P_{data}$. This architecture is advantageous for creating arbitrary faces because data generation can be simply done by sampling $z$. Yet, we need to pay the complexity for the face manipulation because we should find $z$ corresponding to any face for the reconstruction/editing. Finding $z$ of arbitrary image is equivalent to conducting the inverse generation process, which is extremely complex and challenging. Although several studies \cite{donahue2016adversarial,dumoulin2016adversarially} have addressed this problem by learning a bidirectional mapping between $P_z$ and $P_{data}$, their generation and reconstruction quality is less attractive than standard GANs. Most recently, high quality bidirectional GAN  \cite{bang2018high} has been developed for learning the fast and accurate inverse generation process. Authors improve the generation and reconstruction quality of existing bidirectional GANs using namely a connection network, which transforms the feature of discriminator to the latent variable $z$. We employ this connection network into our framework for finding $z$, thus reconstruct the input facial image by generating it from its latent vector $z$. In this way, we can bypass the use of encoder-decoder architecture while keeping the ability of recovering the latent vector of any input image. Given the latent vector of input facial image or that of arbitrary facial image, we manipulate its attribute using the feedback from attribute classifier. 

The main contributions of our study can be summarized as follows. 1) Our algorithm can generate realistic arbitrary faces as well as input faces with desirable multi-attributes. 2) Owing to the attractive nature of GAN latent space, we can easily identify a novel attribute subspace by analyzing the GAN latent space. As a result, our model can be used, without re-training, for manipulating various other attributes, which are not used for training the attribute classifier. 3) Our model is more flexible to structural variations in attributes such as poses because image level information is not transferred to the output. (e.g., skip connections) 4) We can control the degree of attribute effects without additional training or information.

\section{Related Work}
\subsection{Conditional GAN}
While most of generative models have developed the explicit model distribution for learning the data distribution \cite{goodfellow2016nips}, generative adversarial networks (GAN) proposes an implicit approach to learning the data distribution without model assumption or variational bounds; training the image generation process until the distribution of generated images (i.e., model distribution) can approach to that of real images (i.e., data distribution) in the Jensen-Shannon distance sense. However, training GAN is notoriously unstable because the formulation requires finding Nash equilibrium that is tricky to find with gradient descent method \cite{salimans2016improved}. Lately, the training instability has been addressed by various algorithms by developing the stable network architecture \cite{radford2015unsupervised} or introducing robust metrics  \cite{arjovsky2017wasserstein,kodali2018convergence}. To achieve the stability and high quality of GAN, various studies suggest additional information for better posing the generation problem. In fact, from the aspects of application scenario, it is important to control the generation process as we intend, instead of generating the random images. 

To this end, conditional GAN \cite{mirza2014conditional} suggests a new framework to control the semantics of generated samples; they formulate the problem as reproducing the conditional data distribution by training the conditional model distribution. Specifically, authors utilize a semantic vector as a condition of both the generator and discriminator. Unfortunately, upon the complexity of semantic information, the complexity of data distribution is also increased, thus the discriminator is overloaded \cite{chongxuan2017triple}. To alleviate this additional burden, Chongxuan et al. and Odena et al. \cite{chongxuan2017triple,odena2016conditional} propose the separate classification module for handling the semantic information so that the discriminator is focused on evaluating the model distribution. Their architectures impose a cross entropy loss from the classifier to the generator, that guides the conditional generation task. In this work, we adopt a multi-label classifier that decomposes the attribute learning from the discriminator.

\subsection{Facial Attribute and Generation}
The work aiming to edit facial attribute targeted to modify one attribute for one training \cite{shen2017learning,kaneko2017generative}. Shen et al. \cite{shen2017learning} conducts editing the facial attribute by learn the region where the target attribute exists on the face images. 
He et al. \cite{he2017arbitrary} adopts encoder-decoder and classifier. By using skip connection between encoder and decoder, they enhance the reconstruction quality. Choi et al. \cite{choi2017stargan} adds cycle consistency loss to generator's loss to reconstruct images. The entire architecture is similar to multi-cycle-gan with classifier. He et al. \cite{he2017arbitrary} and Choi et al. \cite{choi2017stargan} succeed to edit the attributes of facial images but they can not generate images because of meaningless latent vectors and requiring the input images. SLGAN \cite{yin2017semi} introduces themselves as the first work aiming both the semantic controlled generation and attribute editing. Although they can do both of them, the quality of results is not so good to use as an application. Nevertheless the limitation on quality, the significance is on conducting both generation and editing attributes with one training procedure. In this work, we focus on both facial image generation having desired attributes and editing facial attributes with one training procedure and without degradation on quality of generated or modified samples and stability of training.



\section{Background}
\subsection{Connection Network}
Connection network \cite{bang2018high} serves to estimate a latent vector from a feature vector, which is extracted by GAN discriminator for distinguishing the real and fake. By utilizing this connection network, it is possible to estimate the latent vector of an arbitrary image, either fake or real. Then, the estimated latent vector can be input to a generator to restore the corresponding image. As a result, we establish the bidirectional mapping between the latent vector and image; the generator maps from the latent to the image and the connection network maps from the image to the latent vector. 

Unlike the traditional approaches to developing bidirectional GANs, connection network defines the mapping relationship in low dimensional spaces, low dimensional feature vector and low dimensional latent vector, thus efficient for training. Moreover, because the generator is not involved in estimating the latent vector, the model capacity of generator is solely utilized for improving the image generation quality. 



\subsection{Selective Learning for Classification} \label{AttCNN}
To govern the effect of attributes in image generation, it is necessary to deliver attribute information to the generator network. Existing attribute editing techniques achieve this objective by adapting the attribute classifier \cite{odena2016conditional,he2017arbitrary}. 

Similar to existing techniques, we also aim to govern facial attributes of the resultant faces using the attribute classifier. To develop the highly accurate classifier, it is important to have a balanced database. Unfortunately, real-world database including CelebA \cite{liu2015faceattributes} shows the significant bias among different attributes. Several studies \cite{rudd2016moon,hand2018doing} have developed to alleviate bias in data distribution. For example, Hand et al. \cite{hand2018doing} suggest balancing the number of attribute samples within each batch. Their approach did not require additional network or external database. Instead, they first set the occurrence ratio of each attribute, and select training samples for each batch according to this ratio. This scheme balances the training data by 1) relatively ignoring the attribute samples that exceed the target ratio, and 2) assigning the larger weight for the insufficient amount of attribute samples.  

Although this idea of selective learning is simple and effective, data selection inherently prevents from fully utilizing the entire dataset; it intentionally drops samples from the category with many training data. In this work, we fully utilize the entire training dataset by introducing the weight regularization for both the large and small amount of attribute samples, instead of data selection.   
\section{Editable Generative Adversarial Networks}
The major architectural difference of our model compared to previous models is that we do not use the encoder to transform the image into the latent vector for reconstruction. Instead, we directly generate the image from the latent vector, which is estimated by the separate module, the connection network. For this reason, our networks are capable of generating images and editing attributes simultaneously.  
\subsection{Formulation}
Our network model consists of four networks including a generator $G$, a discriminator $D$, an attribute classifier $C$, and a connection network $C_n$. First, the generator maps the randomly sampled latent vector into the image. The discriminator and classifier have the sample from the generator as the input, and then output the probability of being real or fake and that of presenting attributes, respectively. Meanwhile, the connection network transforms the feature vector extracted by both the discriminator and the classifier to the latent vector. This latent vector rather inputs to the generator for reconstruction/editing. 

The connection network plays a key role in editing attributes of the input image. In our framework, the input image is directly applied to the discriminator and classifier, and results in the feature vector. Then, the connection network translates it to the latent vector corresponding to the input. After that, the generator has the latent vector and a user-specific attribute vector as the input, and  generates the input image, which presents specified attributes assigned by the attribute vector. This procedure is equivalent to modify the attributes of input image. 

\begin{figure}[t!]
    \centering
    \includegraphics[width=0.8\textwidth]{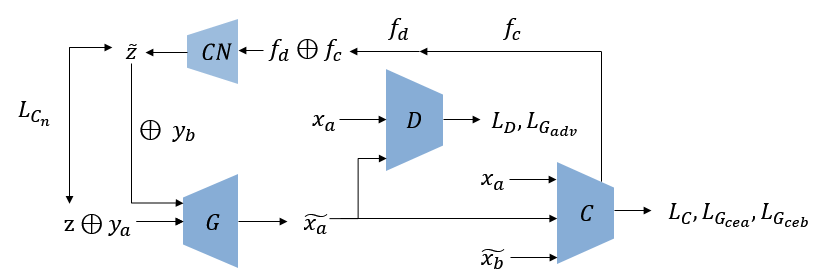}
    \caption{Network architecture of proposed model. $\oplus$ means concatenation along the last dimension. $\tilde{x}$ and $\tilde{z}$ mean generated image and generated latent vector, respectively}
    \vskip -0.5cm
    \label{fig:fig1_structure}
\end{figure}

The novel face generation is identical to the generation process of the standard GANs. That is, we sample a random latent vector from the uniform distribution and set its binary attribute vector. Then, the generator has the latent vector with the attribute vector for producing the novel face image with specified attributes. By changing the attribute vector, it is possible to control the attribute of generated faces. Fig.~\ref{fig:fig1_structure} visualizes the architecture of Editable GANs. 

\noindent\textbf{Adversarial Loss.} 
First, a discriminator provides the critical feedback to a generator for improving the image generation quality. Because the discriminator aims to distinguish real images from dataset and fake images from the generator, this evaluation by the discriminator encourages the generator reproducing the original data distribution. We modify a conditional GAN \cite{mirza2014conditional} to reflect the attribute information in a way that the discriminator becomes irrelevant to the input condition, and formulate the objective for training the discriminator as follows. 
\begin{equation} 
    L_D = \mathbb{E}[\log(D(x))] + \mathbb{E}[\log(1-D(G(z, y_a)))].
\end{equation}
Note $x$ is an image, $z$ is a latent vector, and $y_a$ is the attribute vector of an image ($x$).

\noindent\textbf{Attribute Loss.}
Although the discriminator assesses the general quality of images, it is incapable of evaluating the presence of the desirable attributes in the output image. It is because either the presence or absence of attributes is equally probable, thus does not contribute for the image quality. To ensure the presence of desirable attributes in the resultant image, the attribute classifier is employed. Similar to existing techniques, the attribute vector is a binary multi-dimensional vector, where positive label is $1$, meaning that the attribute appears in the image, and negative label is $0$, meaning that the attribute does not appear in the image. As discussed in Sec.~\ref{AttCNN} celebA dataset shows significant bias among different attribute, the attribute classification should deal with the inbalance of dataset. To alleviate the bias on dataset, we adopt the similar method proposed in \cite{hand2018doing}. Hand et al. \cite{hand2018doing} balances the gradients from positive samples and negative samples for backpropagating at each batch. Specifically, they ignore the gradients from the samples of which attributes exceed a target ratio. Meanwhile, they increase a weight (i.e., greater than 1) for the gradients from the samples of which attributes are smaller than the target ratio. Rather ignoring the the samples, we provide weights to both of exceeding samples and insufficient samples. Let $w_p$ be the weight of positive samples and $w_n$ be that of negative ones. For each attribute, $w_p$ is calculated by dividing the batch size by $2 \times N$. Likewise, $w_n$ for each attribute is computed by dividing the batch size by $2 \times$($M$ - $N$), where $N$ is the total number of labels and $M$ is the batch size.
\begin{equation}
    L_C = \sum_i{-w_{pi} y_i  \log(S(C(x))) - w_{ni}  (1-y_i) \log(1-S(C(x)))}
\end{equation}
Note that $S$ means sigmoid function and $y_i$ means the $i_{th}$ attribute of the input image $x$. $w_p$ and $w_n$ are selective weight of positive label and negative label, respectively.

\noindent\textbf{Training the Generator.} A generator aims 1) to deceive the discriminator by generating realistic faces from the latent vector, and 2) to fool the attribute classifier by producing the desirable attributes in the output image. To this end, both adversarial loss and an attribute constraint are imposed to train the generator.

\begin{equation}
    \begin{aligned}
    &L_{adv\ } = -\mathbb{E}[\log(1-D(G(z, y_a)))]\\
    &L_{G_{ce_a}} = L_C(G(z, y_a))\\
    &L_{G_{ce_{\tilde{a}}}} = L_C(G(\tilde{z}, y_{\tilde{a}}))
    \end{aligned}
\end{equation}
%
$y_a$ is the attribute of real image (given to discriminator and classifier), $y_{\tilde{a}}$ is randomly sampled attribute. $z$ is latent vector randomly sampled from uniform distribution $U[-1,1]$ and $\tilde{z}$ is reconstructed latent vector by connection network.

\noindent\textbf{Connection Network.} 
A connection network plays a role in mapping the image features to the latent vector, which effectively performs the inverse generation process. Because this connection network is trained independently of the generator, it does not introduce additional constraints to the generator training. In this way, we can bypass the disadvantage of encoder-decoder architecture, which overloads the generator training as the range of latent space is enlarged after induced by the encoder. 

While original connection network is built upon the standard GAN architecture with a single generator and a single discriminator, our network model includes the attribute classifier for imposing the desirable attributes in the output image. To correctly reflect those of attribute information in estimating the latent vector, we aggregate the feature vector of the attribute classifier and that of the discriminator by concatenation. More specifically, we derive two feature vectors; $f_d$ from the discriminator and $f_c(G(z,y_a))$ from the classifier, and concatenate them to form the final feature vector. Based on empirical study, we study that the output feature vector of the last fully connected layer is sufficient for mapping to the latent vector. 
\begin{figure}[t!]
    \centering
    \includegraphics[width=\textwidth]{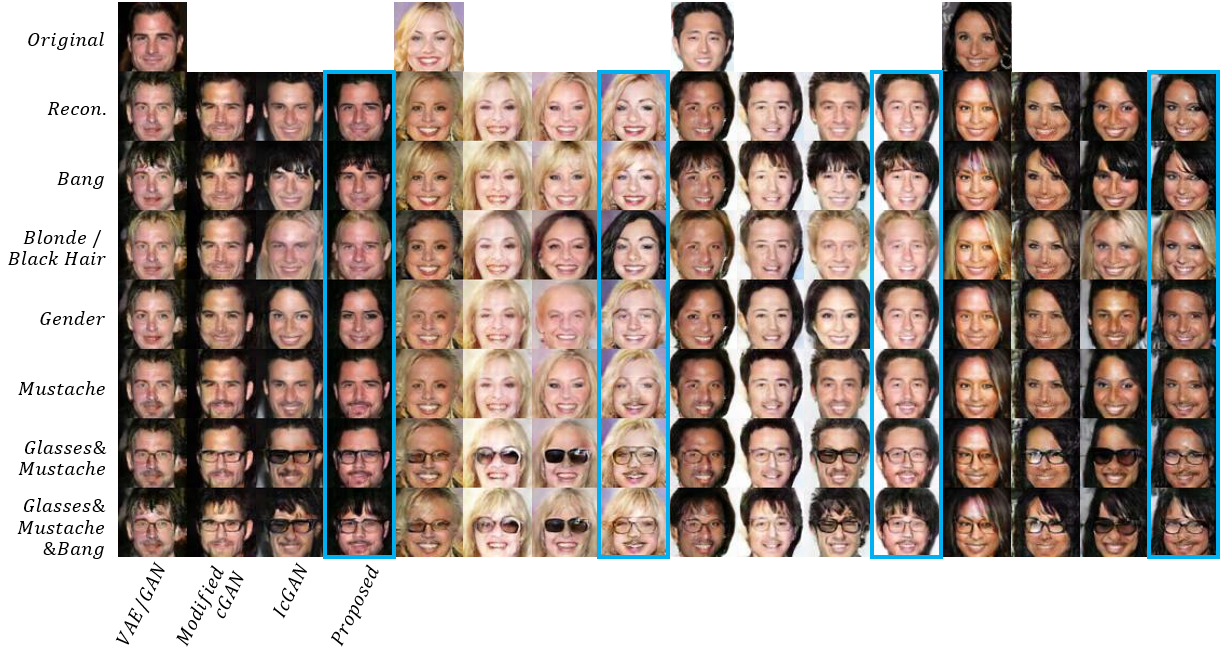}
    \caption{Comparisons of facial attribute editing. The blue box highlights our results. The first three columns are VAE/GAN, modified cGAN, and IcGAN. For each row, the specified attribute(s) is added to the input image}
    \vskip -0.5cm
    \label{fig:fig2_editing}
\end{figure}
By investigating the role of each feature vector, $f_c(G(z,y_a))$ provides the features related to attribute information to the latent vector, thus the reconstructed image from the latent vector is enforced to hold the style of attributes. Meanwhile, $f_d$ maps various elements other than attributes to a unique latent vector. For example, we observe that $f_d$ is associated with identity or structural information. 

\begin{equation}
    \tilde{z} = C_n(f_d(G(z,y_a)) \oplus f_c(G(z,y_a)), y_a)
\end{equation}
\begin{equation}
    L_{C_n} = \mathbb{E}[ \ |z-\tilde{z}| \ ]
\end{equation}
$z$ is the latent vector given to the generator, $\tilde{z}$ is the reconstructed latent vector with the connection network. $y_a$ is the attribute of real image given to discriminator, classifier and generator.
\section{Experiments} 
To evaluate the performance of our Editable GAN, we conduct several tasks on the CelebA dataset \cite{liu2015faceattributes}. CelebA contains 202,599 facial images for training, validation, and testing with the binary labels of 40 attributes. During the experimental study, we choose 10 attributes whose visual features are distinctive from others. We use 192,599 images for training and 10,000 images for test. Under the same configuration, we train three existing algorithms; 1) IcGAN, 2) VAE/GAN and 3) conditional GAN combined with the connection network \cite{bang2018high}, namely the modified cGAN. To implement the modified cGAN, we modify the conditional GAN \cite{mirza2014conditional} by employing the connection network for establishing both face reconstruction and editing. For all experimental studies, the size of images for all experiments is fixed as $64\times64$.
\begin{figure}[t!]
    \centering
            \begin{tabular}{c}
            \includegraphics[width=0.99\textwidth]{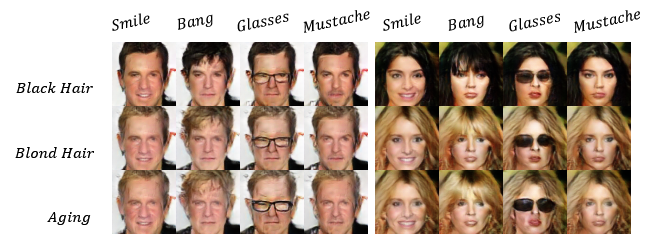} \\
            \text{\ \ \ \ \ \ \ \ \ \ \ IcGAN} \\
            \includegraphics[width=0.99\textwidth]{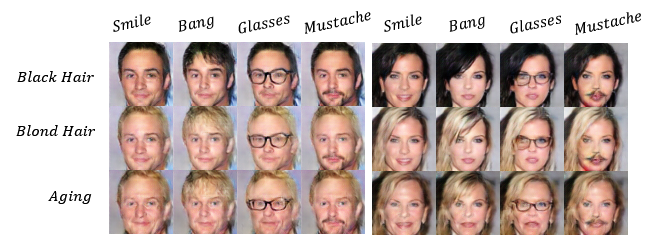} \\
            \text{\ \ \ \ \ \ \ \ \ \ \ \ Proposed} \\
            \end{tabular}
    \caption{Novel image generation with diverse attributes using proposed model and IcGAN. For each algorithm, a young male and young female are drawn by sampling random latent vectors. Then, various attributes are added for visualizing their effects}
    \vskip -0.5cm
    \label{fig:fig3_vsicgan}
\end{figure}    
\subsection{Qualitative Evaluation}
In this paper, our goal is to generate facial images with desired attributes and to edit the attribute of input images. Our evaluation is divided into three folds. 1) To assess the attribute editing of the proposed model and three baseline algorithms (i.e., IcGAN, VAE/GAN and conditional GAN with the connection network), we qualitatively compare the editing results. 2) The quality of novel facial images with desired attributes is evaluated using the proposed model and IcGAN, which produces the most realistic results among three baselines. 3) To show whether the latent space induced by each model is semantically meaningful or now, we perform the latent space interpolation between two latent vectors with interpolated attribute vectors of two images. 4) To show the flexibility of our model to handling structural attribute, we compare our model to AttGAN \cite{he2017arbitrary}. Note that AttGAN focuses on editing attribute, thus their results should be better than any algorithms that perform both image generation and editing. Although their work is optimized for image editing, we show that their framework has the limitation in structural editing, such as poses.  

\begin{figure}[t!]
    \centering
            \begin{tabular}{r c}
            Smile &\includegraphics[align=c, width=0.8\textwidth]{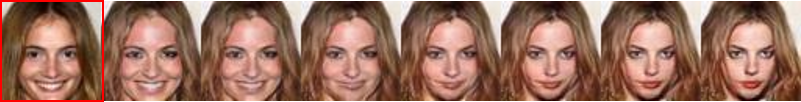} \\
             &\ \ \ \ \ \ \ \ \ \ \ 1 $\xleftrightarrow{{\ \ \ \ \ \ \ \ \ \ \ \ \ \ \ \ \ \ \ \ \ \ \ \ \ \ \ \ \ \ \ \ \ \ \ \ \ \ \ \ \ \ \ \ \ \ \ \ \ \ \ \ \ \ \ \ \ \ \ \ \ \ \ }}$ -1\vspace{1mm}\\
            Mustache &\includegraphics[align=c, width=0.8\textwidth]{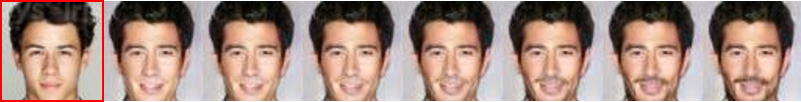}\vspace{0.5mm}\\
            Black hair &\includegraphics[align=c, width=0.8\textwidth]{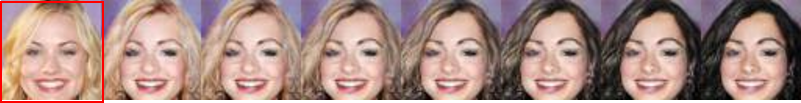}\vspace{0.5mm}\\
            Male &\includegraphics[align=c, width=0.8\textwidth]{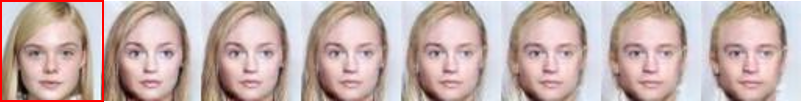}\\
            &\ \ \ \ \ \ \ \ \ \ \ 0 $\xleftrightarrow{{\ \ \ \ \ \ \ \ \ \ \ \ \ \ \ \ \ \ \ \ \ \ \ \ \ \ \ \ \ \ \ \ \ \ \ \ \ \ \ \ \ \ \ \ \ \ \ \ \ \ \ \ \ \ \ \ \ \ \ \ \ \ \ \ }}$ 1\\
            \end{tabular}
    \caption{Controlling the strength of attribute effect. The first row indicates smile attribute between $1$ and $-1$. From the second to forth row, each attribute changes from $0$ to $1$ when their attributes are mustache, black hair and male. For the third row, blond hair varies from $1$ to $0$ at the same time. The input images are marked by the red box}
    \vskip -0.5cm
    \label{fig:fig_4_attribute_interpolation}
\end{figure}    
\noindent\textbf{Facial Attribute Editing.} We choose the input images from a test set, excluding training and validation set. Fig.~\ref{fig:fig2_editing} shows the results of attribute editing from the VAE/GAN, the modified cGAN, IcGAN, and the proposed algorithms with the input image. Focusing on the reconstruction results, it is clear that our reconstruction better copes the identity and style much better than others. Analyzing the results of attribute editing, we observe that modified cGAN occasionally misses the attribute. (e.g., blond or black hair) This is expected because the generator was not penalized due to wrong attribute generation during training. Also, VAE/GAN and IcGAN are incapable of generating the mustache on the female faces. These results indicates that VAE/GAN and IcGAN do not properly disentangle the attribute space. Among three baseline algorithms, IcGAN generally performs better than other two algorithms. However, even IcGAN sometimes introduces undesirable changes; for example, the hairline of forehead from 'Gender' editing is different from that from input faces. Among all algorithms, the proposed model is the most faithful to reflect the facial attributes than others. Additionally, our attribute editing does not modify other unique attributes of the input face. 
Also, our results are more natural and realistic under multiple attribute editing than others. \\
\begin{figure}[t!]
    \centering
        \begin{tabular}{c c c}
        \includegraphics[height=2cm]{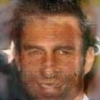} 
        &\includegraphics[height=2cm]{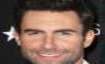}
        &\includegraphics[height=2cm]{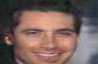} \\
        AttGAN & Input & Proposed \\
        \end{tabular}
    \caption{Editing the orientation of the face. We estimate the latent vector of orientation by subtracting profile face from the frontal face, which is computed by averaging the latent vectors of the same person}
    \label{fig:fig5_vs_att_orient}
    \vskip -0.5cm
\end{figure}   
\noindent\textbf{Generating Facial Image with Desired Attributes.}
Our Editable GAN can conduct the image generation task with desirable attributes by adopting the attribute classifier. Our model is compared with IcGAN that shows best performance among three competitors. In this experiment, multiple attributes are combined for image generation, where attribute information is noted in each column and row. As shown in Fig.~\ref{fig:fig3_vsicgan}, our model can reflect the effect of each attribute and also successfully combine multiple attributes. Moreover, upon varying attributes, our results preserve the identity. Similar to the results of Fig.~\ref{fig:fig2_editing}, IcGAN cannot generate the female face with mustache because IcGAN is vulnerable to decoupling highly correlated attributes.\\
\noindent\textbf{Latent Interpolation of Two Faces and Attributes.}
By performing the latent interpolation between two faces, we show that 1) our results are not accomplished by data memorization, and 2) our latent and attribute spaces are semantically meaningful. During this empirical study, we find that the smooth transition between the attribute vector can induce the semantically natural and meaningful changes in the images across the attribute. While the semantic relationship between adjacent images is varied smoothly, each generated image is still sharp and realistic. 
Because our latent and attribute spaces are semantically meaningful, our model can represent the strength of attribute effect by controlling the intensity of attribute vector. This is interesting because the attribute vector is a binary vector during training. Although our network never observe the degree of attribute effect during training, it automatically interprets the strength of attributes from the intensity of attribute vector. For example, in the first row of Fig.~\ref{fig:fig_4_attribute_interpolation}, 
we apply 'Smile' by interpolating it from $1$ to $-1$, and observe that the changes are smoothly reflected, from a big smile to a cold face. In addition, our model interprets the negative value in the attribute vector, producing a semantically opposite attribute. Again, the negative value is never supervised during training, but negative 'Smile' is translated to the cold face. For 'Mustache' attribute, the amount of hairs increases as it changes to a higher value. 'Black hair' attribute also shows the interesting behavior when manipulating its intensity. In the third row of Fig.~\ref{fig:fig_4_attribute_interpolation}, the input face has the blond hair. Then, we gradually increase 'Black hair' attribute from $0$ to $1$. Focusing on the middle of the interpolation, we can observe that the generated hair color is neither black nor blond, but a blend of two colors. Utilizing the semantic interpretation of our model, we can create diverse variants of a single face.\\
\noindent\textbf{Pose Editing by Latent Space Walking.}
Analogous to the latent space walking, we effectively induce the structural modification such as poses. To do this, we interpolate between one image and its vertically flipped version with a fixed attribute on the latent vector space. Although AttGAN \cite{he2017arbitrary} is not our competitor because they only focus on attribute editing, we compare AttGAN with ours in terms of pose editing by latent space walking. In the general scenario of attribute editing, AttGAN should always perform better than other algorithms to generate and edit faces simultaneously. However, when handling the structural changes, we observe that our model can perform clearly better than AttGAN as shown in  Fig.~\ref{fig:fig5_vs_att_orient}. Although the pose variation has never taught during training, our model can disentangle the pose variation from the facial identity, successfully conducting the pose changes by latent space arithmetic. On the other hand, AttGAN fails to disentangle the pose attribute in latent space, thus incapable of producing the sharp images with smooth pose transition. This failure of structural changes with AttGAN is anticipated by their encoder-decoder architecture and skip connections implemented in their network. Unlike the standard GAN architecture, the latent distribution $P_z$ in the encoder-decoder based models tends to be sparse; its range is typically greater than the range of standard GAN. Hence, the latent space induced by the encoder-decoder architecture rarely possess semantic interpretation. Also, while skip connections are quite effective to accurately reconstruct the input face, they weaken the semantic power of latent space, as discussed in \cite{zhang2017style}. As a result, the operation in the latent space does not yield the meaningful interpretation. Fig.~\ref{fig:fig6_vs_att_inter} shows the result of latent interpolation between one image and its flip. In this experiment, we can see the gradual change of the same type of the experiment shown in Fig.~\ref{fig:fig5_vs_att_orient}. As visualized in Fig.~\ref{fig:fig6_vs_att_inter}, the results from AttGAN is akin to the result of an image interpolation, shown in first row of the same figure. Although each image is produced by the generator after the latent interpolation, the generated image is no longer realistic as if it is an overlap of two images. 
\subsection{Quantitative Evaluation}\label{Quantitative}
For quantitative evaluation, we divide our task into the novel image generation, faithful image reconstruction, and modification. For evaluating the image generation quality, we employ Fréchet Inception Distance (FID) \cite{heusel2017gans}. For measuring the image reconstruction accuracy, we use the structural similarity index (SSIM), the peak signal-to-noise ratio (PSNR). Note that the accurate image reconstruction is critical for achieving successful attribute editing because our attribute editing is based on the image reconstruction framework. Finally, we evaluate the attribute editing by measuring the attribute classification accuracy for the edited images.\\
\begin{figure}[t!]
    \centering
         \includegraphics[align=c,width=\textwidth]{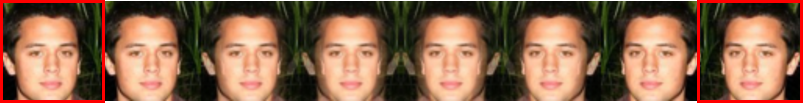}\vspace{1mm}\\
        Image interpolation \\
        \includegraphics[align=c,width=\textwidth]{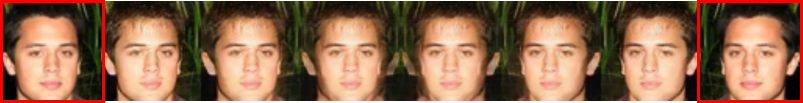}\vspace{1mm}\\
        AttGAN \\
        \includegraphics[align=c,width=\textwidth]{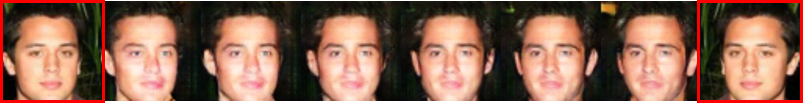}\\
        Proposed\\
    \caption{Interpolation of the face and its vertical flipped on latent vector space. First row shows the interpolation on the image space, weighted summation of rightmost and leftmost. The second and bottom row are done by AttGAN and our model, respectively. The rightmost and leftmost image are input images ({\it highlighted by red box})}
    \label{fig:fig6_vs_att_inter}
    \vskip -0.2cm
\end{figure}
\begin{table}[t!]
\caption{ Comparison of image generation quality. We compare the quality of generated images by FID score (mean and standard deviation) with recent algorithms achieving the generation and attribute editing simultaneously. Note that modified cGAN is implemented by a condtional DCGAN combined with a connection network. The lower the value, the better the quality.}
\centering
\label{tab:FID}
\begin{tabularx}{\textwidth}{@{}l *5{>{\centering\arraybackslash}X}@{}}
\hline
Metric  &VAE/GAN  &cGAN &IcGAN &Proposed \\ \hline
FID     &24.05$\pm$0.33 &23.27$\pm$0.46 &22.86$\pm$0.80 &19.92$\pm$0.73\\ \hline
\end{tabularx}
\end{table}
\textbf{Generation Performance.} 
For assessing the image generation quality, we compare our model with VAE/GAN, modified cGAN, and IcGAN using FID score. For the fair comparison, we repeatedly conduct the experiment 10 times, and report the average and standard deviation of FID score for each model as summarized in Table.~\ref{tab:FID}. Considering that the smaller value for FID stands for the higher quality, our Editable GAN outperforms other competitors; the performance gap is greater than three standard deviations.\\   
\begin{table}[t!]
\caption{Comparison of reconstruction performance in terms of SSIM and PSNR. (mean and standard deviation) The higher the value, the better the quality.}
\centering
\label{tab:SSIM}
\begin{tabularx}{\textwidth}{@{}l *5{>{\centering\arraybackslash}X}@{}}
\hline
Metric  &VAE/GAN  &cGAN &IcGAN &Proposed \\ \hline
SSIM    &0.39$\pm$0.0200   &0.50$\pm$0.0081   &0.46$\pm$0.0092   &0.52$\pm$0.0088\\ \hline
PSNR    &12.98$\pm$0.19   &16.15$\pm$0.20   &15.19$\pm$0.19   &16.54$\pm$0.22\\\hline
\end{tabularx}
\vskip -0.5cm
\end{table}
\textbf{Reconstruction Performance.}
For successful attribute editing, it is important to confirm whether the input face can be faithfully reconstructed using each model. For that, we evaluate our reconstruction accuracy by SSIM and PSNR using test images of CelebA dataset, and report the average and standard deviation for VAE/GAN, Modified cGAN, IcGAN, and our Editable GAN. 
Table.~\ref{tab:SSIM} summarizes the accuracy of each model. Among all, our model outperforms all others in terms of SSIM and PSNR. It is important to stress that the performance gap between ours and any existing model is greater than two standard deviations. From this result, we can conclude that our model possesses the greater representation power than other competitors.\\  
\begin{figure}[t!]
    \centering
    \includegraphics[width=\textwidth]{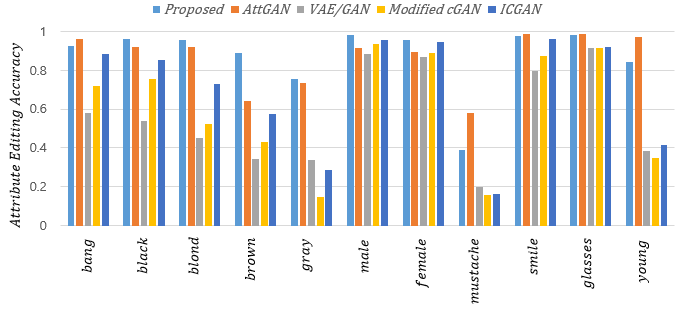}
    \vskip -0.1cm
    \caption{Facial attribute editing accuracy of each model}
    \label{fig:graph_cls}
    \vskip -0.5cm    
\end{figure}
\textbf{Editing Performance.}\label{EditingPerformance}
To evaluate the attribute modification quality, it is necessary to check whether the effect of desirable attributes is correctly reflected after image editing. To this end, we develop the pretrained attribute classifier, and use it as the evaluation classifier for measuring how well the attribute is recognizable. To improve the performance of the evaluation classifier, we include several residual blocks \cite{he2016deep} and apply selective learning method proposed in \cite{hand2018doing}. We test this evaluation classifier and confirm that its average accuracy is 93.22\% over 10 attributes. More specifically, training dataset of CelebA for 10 attributes is applied for developing the evaluation classifier. During this experiment, we exam a single attribute modification case for all models, and the attribute classification accuracy is visualized in Fig.~\ref{fig:graph_cls}. For the comparison, we evaluate the attribute classification accuracy of VAE/GAN, modified cGAN, IcGAN, AttGAN, and our Editable GAN. Based on these experimental results, we observe that the proposed model outperforms our competitors; VAE/GAN, modfied cGAN, and IcGAN. 
Compared to AttGAN \cite{he2017arbitrary}, we expect that AttGAN should always show the higher performance because their model focuses on attribute editing, specialized for improving the editing performance. Interestingly, Editable GAN is compatible with AttGAN in attribute editing. For six out of 10 attributes (e.g., black, blond, brown, gray, male, and female), the proposed model even outperforms AttGAN. 
Based on three quantitative comparisons, we can conclude that our Editable GAN is an effective solution for simultaneously generating and editing faces with desired attributes. 
\section{Conclusion}
This paper introduces an Editable Generative Adversarial Network (Editable GAN), which establishes 1) the multiple attribute editing of the input face and 2) the novel face generation by controlling semantic attributes. For attribute editing, our model utilizes the attribute classifier to modify multiple attributes of the input face. For the face reconstruction and generation perspective, the proposed model adopts the connection network \cite{bang2018high} that estimates the latent variable of the input. We separate the training of connection network from GAN training, thus stabilize the entire training process and retain the quality of image generation. Furthermore, because the image level information is not utilized during training, our model can edit the structural variations such as poses. More importantly, the proposed model is flexible to model the strength of attribute effects, and to handle other semantic attributes, not even observed during training. Owing to the semantic interpretation in the latent space, we successfully disentangle the attributes of unseen categories, and utilize them for image editing. 

\bibliographystyle{splncs}

\begin{thebibliography}{10}
\bibitem{hand2018doing}
Hand, E.M., Castillo, C., Chellappa, R.:
\newblock Doing the best we can with what we have: Multi-label balancing with
  selective learning for attribute prediction, AAAI (2018)
\end{thebibliography}


\end{document}